\documentclass{vllm}

\usepackage[margin=1.2in]{geometry}
\usepackage{amsmath,amssymb,amsthm}
\usepackage{inconsolata}
\usepackage{listings}
\usepackage{pifont}
\usepackage{enumitem}
\usepackage{tabularx}
\usepackage{tikz}
\usepackage{algorithm}
\usepackage{algorithmic}
\usepackage[round]{natbib}
\usetikzlibrary{arrows.meta,calc,fit,positioning,shapes.geometric,
                backgrounds,decorations.pathreplacing}

\definecolor{sigblue}{HTML}{D6E4F0}
\definecolor{sigbluebdr}{HTML}{4A7FB5}
\definecolor{projgreen}{HTML}{D9EDDB}
\definecolor{projgreenbdr}{HTML}{4A9B5A}
\definecolor{decamber}{HTML}{FDE8CD}
\definecolor{decamberbdr}{HTML}{C8862A}
\definecolor{priorose}{HTML}{F2D5D5}
\definecolor{priorosebdr}{HTML}{B04A4A}
\definecolor{loopteal}{HTML}{D2ECE9}
\definecolor{looptealbdr}{HTML}{2D8C82}
\definecolor{liveorange}{HTML}{FFF0DB}
\definecolor{liveorangebdr}{HTML}{D98C21}
\definecolor{cachgreen}{HTML}{E6F4E6}
\definecolor{cachgreenbdr}{HTML}{4A9B5A}
\definecolor{fwpurple}{HTML}{E8DFF0}
\definecolor{fwpurplebdr}{HTML}{7B5EA7}

\definecolor{polcyan}{HTML}{D4EEF6}
\definecolor{polcyanbdr}{HTML}{2788A5}
\definecolor{enfgray}{HTML}{E5E5E5}
\definecolor{enfgraybdr}{HTML}{666666}

\definecolor{yamlkey}{HTML}{2D5F8A}
\definecolor{yamlval}{HTML}{6A8759}
\definecolor{yamlcmt}{HTML}{888888}
\definecolor{pykey}{HTML}{CF222E}
\definecolor{pystr}{HTML}{0A3069}
\definecolor{pycmt}{HTML}{6E7781}

\lstdefinestyle{yaml}{
  basicstyle=\ttfamily\scriptsize,
  columns=fullflexible,
  keepspaces=true,
  showstringspaces=false,
  commentstyle=\color{yamlcmt}\itshape,
  keywordstyle=\color{yamlkey}\bfseries,
  stringstyle=\color{yamlval},
  morecomment=[l]{\#},
  frame=single,
  framerule=0.4pt,
  rulecolor=\color{black!25},
  backgroundcolor=\color{black!2},
  aboveskip=4pt,
  belowskip=4pt,
  xleftmargin=3pt,
  xrightmargin=3pt,
  escapeinside={(*@}{@*)},
}

\lstdefinestyle{dsl}{
  basicstyle=\ttfamily\scriptsize,
  columns=fullflexible,
  keepspaces=true,
  showstringspaces=false,
  keywordstyle=\color{yamlkey}\bfseries,
  stringstyle=\color{yamlval},
  commentstyle=\color{yamlcmt}\itshape,
  morecomment=[l]{\#},
  morekeywords={SIGNAL,ROUTE,BACKEND,GLOBAL,PLUGIN,DECISION_TREE,
                SIGNAL_GROUP,TEST,NETWORK,DEPLOY,AGENT,WHEN,
                PRIORITY,MODEL,ACTION,IF,ELSE,AND,OR,NOT},
  frame=single,
  framerule=0.4pt,
  rulecolor=\color{black!25},
  backgroundcolor=\color{black!2},
  aboveskip=4pt,
  belowskip=4pt,
  xleftmargin=3pt,
  xrightmargin=3pt,
}

\lstdefinestyle{python}{
  language=Python,
  basicstyle=\ttfamily\scriptsize,
  columns=fullflexible,
  keepspaces=true,
  showstringspaces=false,
  keywordstyle=\color{pykey}\bfseries,
  stringstyle=\color{pystr},
  commentstyle=\color{pycmt}\itshape,
  frame=single,
  framerule=0.4pt,
  rulecolor=\color{black!25},
  backgroundcolor=\color{black!2},
  aboveskip=4pt,
  belowskip=4pt,
  xleftmargin=3pt,
  xrightmargin=3pt,
}

\lstdefinestyle{xml}{
  language=XML,
  basicstyle=\ttfamily\scriptsize,
  columns=fullflexible,
  keepspaces=true,
  showstringspaces=false,
  keywordstyle=\color{pykey}\bfseries,
  stringstyle=\color{pystr},
  commentstyle=\color{pycmt}\itshape,
  frame=single,
  framerule=0.4pt,
  rulecolor=\color{black!25},
  backgroundcolor=\color{black!2},
  aboveskip=4pt,
  belowskip=4pt,
  xleftmargin=3pt,
  xrightmargin=3pt,
  morestring=[b]",
  morekeywords={rpc,edit-config,target,candidate,config,
    policy,version,source-hash,signals,signal,name,kind,
    threshold,model,candidates,routing,decision-tree,
    branch,priority,condition,backend,network,
    network-endpoint,host,port,skill,pii-types-allowed},
}

\newcolumntype{Y}{>{\raggedright\arraybackslash}X}

\title{\textbf{From Inference Routing}\\[2pt]
       \textbf{to Agent Orchestration}\\[4pt]
       {\large Declarative Policy Compilation\\
        with Cross-Layer Verification}}

\author{%
  Huamin Chen$^{1}$ \quad
  Xunzhuo Liu$^{1}$ \quad
  Bowei He$^{2,*}$ \quad
  Xue Liu$^{1,2,3,4}$
}

\affiliation[1]{vLLM Semantic Router Project}
\affiliation[2]{MBZUAI}
\affiliation[3]{McGill University}
\affiliation[4]{Mila}
\affiliation[*]{Corresponding author}

\date{March 2026}

\abstract{%
The Semantic Router DSL is a non-Turing-complete policy
language deployed in production for per-request LLM
inference routing: content signals (embedding similarity,
PII detection, jailbreak scoring) feed into weighted
projections and priority-ordered decision trees that
select a model, enforce privacy policies, and produce
structured audit traces---all from a single declarative
source file.  Prior work has established conflict-free
compilation for probabilistic predicates and positioned
the DSL within the Workload--Router--Pool inference
architecture.

This paper extends the same language from stateless,
per-request routing to stateful, multi-step agent
workflows---the full path from inference gateway to
agent orchestration to infrastructure deployment.
We propose \emph{declarative policy compilation}: the
DSL compiler emits verified decision nodes for
orchestration frameworks (LangGraph), Kubernetes
artifacts (NetworkPolicy, Sandbox CRD, ConfigMap),
YANG/NETCONF payloads, and protocol-boundary gates
(MCP, A2A)---all from the same \texttt{.sr} source.

Because the language is non-Turing-complete, the
compiler adds guarantees that complement each target's
own checks: exhaustive routing, conflict-free branching,
referential integrity across deployment artifacts, and
audit traces structurally coupled to the decision logic.
Because signal definitions are shared across all targets,
a threshold change propagates from inference gateway to
agent gate to infrastructure artifact in one compilation
step---eliminating the cross-team coordination that is
the primary source of policy drift.
We ground the approach in four pillars---auditability,
cost efficiency, verifiability, and tunability---and
identify the precise verification boundary at each
layer.  A critical analysis examines each claim,
identifies necessary caveats, and characterizes the
conditions under which the unified policy approach
provides the most value.
}

\begin{document}
\maketitle

\section{Introduction}
\label{sec:intro}

The Semantic Router DSL is a non-Turing-complete policy
language that governs per-request LLM inference routing
in production.  A routing policy declares content
\textsc{Signal}s---embedding similarity, PII classifiers,
jailbreak detectors, keyword matchers---that feed into
weighted \emph{projections} (composite scores mapped to
discrete bands via thresholds), which in turn feed into
priority-ordered \emph{decision trees} that select a
model backend.  The compiler emits a YAML routing
configuration consumed by the inference gateway, which
evaluates the policy on every request, producing a fully
traceable routing decision before any model call occurs
\citep{chen2026tdn}.

Three prior papers establish the DSL's foundations.
\emph{Transparent Discrete Networks} \citep{chen2026tdn}
shows that the DSL's signals--projections--decisions
pipeline is isomorphic to a discrete feedforward network
whose parameters are analytically optimizable from
execution traces---tuning is symbolic program
transformation, not gradient descent.
\emph{ProbPol} \citep{liu2026conflictfree} formalizes
conflict detection for policies whose conditions are
probabilistic ML signals, introducing a three-level
decidability hierarchy and Voronoi normalization via
temperature-scaled softmax to eliminate co-firing.
The \emph{Workload--Router--Pool} architecture
\citep[WRP;][]{chen2026wrp} positions the DSL as the
Router pillar's policy language within the full LLM
inference optimization space.

All three papers treat the DSL as a \emph{stateless,
per-request decision function}: a request arrives, signals
fire, a model is selected, and the decision is traced.
Yet LLM-powered applications are increasingly
\emph{multi-step agent workflows}: a user request
triggers a chain of model calls, tool invocations via
the Model Context Protocol
\citep[MCP;][]{mcp2024}, delegations to other agents via
Agent2Agent \citep[A2A;][]{a2a2025}, and external API
calls---each step carrying its own cost, latency, safety,
and compliance implications.  Frameworks such as LangGraph
\citep{langgraph2024}, Temporal \citep{temporal2024}, and
CrewAI \citep{crewai2024} orchestrate these workflows by
managing state, control flow, and fault tolerance.

These orchestration frameworks excel at their job---state
management, fault tolerance, control flow---but
\emph{static analysis of the policy decisions embedded
in each step} (which model to call, whether a tool
invocation is safe, whether routing logic is exhaustive)
is orthogonal to orchestration and currently left to
ad-hoc Python functions.

This paper asks: \textbf{can the same DSL that already
governs per-request inference routing extend to
multi-step agent workflows, and if so, what does a
unified policy language across the full stack provide
that separate, ad-hoc implementations cannot?}

The answer is a \emph{multi-target compiler}.  The DSL
remains a policy language---it evaluates signals and
returns decisions---while orchestration frameworks
handle step sequencing, retries, and state.  What changes
is the \emph{compilation target}: the same \texttt{.sr}
source that today emits YAML routing configuration for
the inference gateway now also emits verified LangGraph
decision nodes, OpenClaw gateway policy bundles,
Kubernetes deployment artifacts (NetworkPolicy, Sandbox
CRD, ConfigMap), YANG/NETCONF payloads, and
protocol-boundary gates for MCP and A2A.
A threshold change in the DSL source propagates to every
target on recompilation---the property that is hardest to
maintain when each layer is coded independently.

\begin{figure}[t]
  \centering
  \resizebox{\columnwidth}{!}{%
  \begin{tikzpicture}[
      layer/.style={draw=#1!70, very thick, rounded corners=3pt,
                    minimum height=0.85cm, inner sep=4pt,
                    fill=#1!18, font=\scriptsize,
                    text width=5.0cm, align=left},
      arr/.style={-{Latex[length=2mm]}, line width=0.8pt,
                  draw=black!40},
    ]
    % Define Y positions first
    \def\yA{1.65}   % inference
    \def\yB{0.55}   % orchestration
    \def\yC{-0.55}  % protocols
    \def\yD{-1.65}  % infrastructure
    \def\xL{0}      % DSL source x
    \def\xR{5.0}    % layer boxes x

    % Left: DSL source — tall enough to span all layers
    \node[draw=black!40, very thick, rounded corners=4pt,
          fill=black!4, inner sep=6pt,
          minimum height=4.2cm,
          font=\scriptsize\bfseries, align=center]
      (dsl) at (\xL,0)
      {Single \texttt{.sr} source\\[3pt]
       \tiny\normalfont signals, thresholds,\\
       trees, networks, agents};

    % Right: compilation targets
    \node[layer=decamberbdr] (infer) at (\xR,\yA)
      {\textbf{Inference routing}
       {\tiny (Semantic Router)}\\[-1pt]
       {\tiny YAML config, model selection, trace API}};

    \node[layer=looptealbdr] (agent) at (\xR,\yB)
      {\textbf{Agent orchestration}
       {\tiny (LangGraph / OpenClaw)}\\[-1pt]
       {\tiny Safety gates, intent routing, tool authz}};

    \node[layer=polcyanbdr] (proto) at (\xR,\yC)
      {\textbf{Protocol boundaries}
       {\tiny (MCP / A2A / AP2)}\\[-1pt]
       {\tiny Content-aware gating at call boundaries}};

    \node[layer=enfgraybdr] (infra) at (\xR,\yD)
      {\textbf{Infrastructure}
       {\tiny (Kubernetes / YANG)}\\[-1pt]
       {\tiny NetworkPolicy, Sandbox CRD, NETCONF}};

    % Arrows: purely horizontal lines from DSL east edge
    \foreach \tgt in {infer,agent,proto,infra}
      \draw[arr] (dsl.east |- \tgt) -- (\tgt.west);

    % Status labels: anchored to right edge of each box
    \foreach \tgt/\col/\txt in {
      infer/decamberbdr/deployed,
      agent/looptealbdr/this paper,
      proto/polcyanbdr/this paper,
      infra/enfgraybdr/this paper}
      \node[font=\tiny\sffamily, text=\col, right=4pt]
        at (\tgt.east) {\txt};
  \end{tikzpicture}%
  }
  \caption{End-to-end compilation targets from a single
    DSL source.  Inference routing is the established
    deployment; this paper extends the same language to
    agent orchestration, protocol boundaries, and
    infrastructure artifacts.  Signal definitions,
    thresholds, and verification guarantees are shared
    across all targets; a change in one DSL parameter
    propagates to every layer on recompilation.}
  \label{fig:layers}
\end{figure}
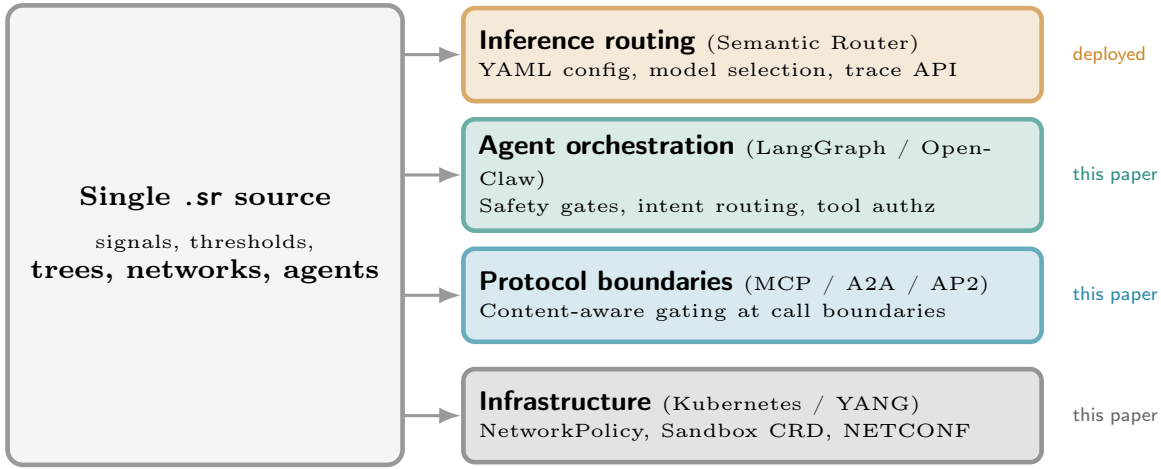

This framing yields four concrete contributions:

\begin{enumerate}[leftmargin=*,itemsep=2pt]
  \item[\textbf{C1.}]
    A \textbf{multi-target compilation} from DSL
    constructs (\textsc{Signal}, \textsc{Decision\_Tree},
    \textsc{Signal\_Group}) to four target layers:
    inference routing, orchestration decision nodes,
    protocol gates, and infrastructure artifacts
    (\S\ref{sec:compilation}).

  \item[\textbf{C2.}]
    A \textbf{four-pillar analysis}---auditability,
    cost efficiency, verifiability, tunability---showing
    that the guarantees established for inference routing
    extend to agent workflows with explicit boundaries
    on what each pillar covers (\S\ref{sec:pillars}).

  \item[\textbf{C3.}]
    A \textbf{complementarity argument} showing how a
    single DSL source adds a consistent policy layer
    across the agent--inference--tool boundary
    (\S\ref{sec:gaps}), with the same signal definitions
    governing all three.

  \item[\textbf{C4.}]
    A \textbf{critical analysis} that examines each claim
    against anticipated objections, identifies necessary
    caveats, and characterizes when the overhead is
    justified (\S\ref{sec:discussion}).
\end{enumerate}

Figure~\ref{fig:layers} summarizes the architecture.
The inference routing layer (top) is the established
deployment; the agent, protocol, and infrastructure
layers are the extensions proposed in this paper.  A
single \texttt{.sr} source file generates artifacts for
all four layers, with cross-layer consistency enforced
by the compiler.

\section{Background}
\label{sec:background}

\subsection{The Semantic Router DSL}

The DSL is a non-Turing-complete policy language with
five core block types: \textsc{Signal} (a probabilistic
ML detector returning a score in $[0,1]$, thresholded
to Boolean), \textsc{Route} (a priority-ordered rule with
a \textsc{When} clause composing signals via
\textsc{And}/\textsc{Or}/\textsc{Not}), \textsc{Backend}
(a model or action target), \textsc{Plugin} (a middleware),
and \textsc{Global} (default settings).  Two additional
constructs address the probabilistic nature of signals:
\textsc{Signal\_Group} (softmax normalization over a set
of signals to guarantee at most one fires) and
\textsc{Decision\_Tree} (a tree of \textsc{If}/\textsc{Else}
branches compiled as a Firewall Decision Diagram).

The language is deliberately constrained: no mutable
state, no side effects, no loops, no recursion, finite
typed signal space.  This restriction is what enables
static analysis.  The design follows the same philosophy
as NetKAT \citep{anderson2014netkat}: a domain-specific
algebra whose restricted expressiveness admits decidable
verification.

The compiler emits multiple targets from a single source.
The established target is YAML routing configuration
consumed by the Semantic Router inference gateway, where
the DSL governs per-request model selection, privacy
routing, and confidence-based model escalation
\citep{chen2026tdn}.  Additional targets include
Kubernetes CRDs and Helm values.  This paper proposes
extending the compiler to emit decision nodes for
orchestration frameworks, protocol-boundary gates, and
YANG/NETCONF payloads---reusing the same signal
definitions and verification guarantees that are already
validated at the inference routing layer.

\subsection{Orchestration Frameworks}
\label{sec:bg:orch}

Agent orchestration takes two architectural forms.
\emph{Graph-based} frameworks such as LangGraph
\citep{langgraph2024}, Temporal \citep{temporal2024},
and CrewAI \citep{crewai2024} model a workflow as a
directed graph of \emph{nodes} (computation steps)
connected by \emph{edges} (control flow).  State flows
through the graph as a typed record; routing decisions
are expressed as \emph{conditional edges}: a
user-defined function inspects the updated state and
selects the next node.
\emph{Gateway-based} platforms such as OpenClaw
\citep{openclaw2025} take a different approach: a
central gateway routes inbound messages to isolated
agents via a configuration-driven binding table, with
tool access, sandboxing, and session policies declared
in JSON configuration and enforced by plugin hooks.
The agent's internal loop (an embedded coding agent
runtime) is LLM-driven, but \emph{host-level} policy---which
tools are available, which channels are accepted, how
sandboxing is applied---is static and declarative.

In both forms, the framework validates its own
structural concerns: graph-based compilers check node
reachability and schema consistency; gateway platforms
validate binding tables and tool profiles.  Semantic
properties of the \emph{policy logic}---exhaustiveness,
conflict-freedom, cross-gate consistency---fall outside
the orchestration layer in both cases.
Table~\ref{tab:lg_gaps} illustrates how the DSL compiler
contributes a complementary set of semantic checks.

\begin{table}[t]
  \centering\small
  \caption{Complementary checks performed by
    orchestration compilers and the DSL compiler.
    Each focuses on its own concern.}
  \label{tab:lg_gaps}
  \begin{tabularx}{\columnwidth}{lcc}
    \toprule
    \textbf{Property} & \textbf{Orchestration} & \textbf{DSL} \\
    \midrule
    \multicolumn{3}{l}{\emph{Graph structure}} \\
    \quad Node connectivity        & \checkmark &  \\
    \quad Edge targets exist       & \checkmark &  \\
    \quad State schema validation  & \checkmark &  \\
    \addlinespace
    \multicolumn{3}{l}{\emph{Routing semantics}} \\
    \quad Routing exhaustiveness   &  & \checkmark \\
    \quad Branch conflict detection &  & \checkmark$^*$ \\
    \quad Dead branch detection    &  & \checkmark$^*$ \\
    \quad Signal co-firing prevention &  & \checkmark \\
    \quad Referential integrity    &  & \checkmark \\
    \quad Cross-artifact consistency &  & \checkmark \\
    \quad Structured audit traces  &  & \checkmark \\
    \bottomrule
    \multicolumn{3}{l}{\scriptsize
      $^*$Decidable for crisp and geometric signals;
      empirical only for classifiers.}
  \end{tabularx}
\end{table}

\subsection{Opportunities for a Policy Layer}

The Semantic Router already provides a full policy layer
for single-request inference: content signals evaluate
the request, projections aggregate signal scores, and a
priority-ordered decision tree selects the model backend
\citep{chen2026tdn}.  In multi-step workflows, the same
policy concerns recur \emph{at every step}---but each
step is currently coded independently.  Three areas rely
on ad-hoc application code where the DSL's established
inference-routing machinery can help:

\begin{enumerate}[leftmargin=*,itemsep=2pt]
  \item \textbf{Agent $\to$ Inference.}
    Agent frameworks typically assign a fixed model per
    node.  The DSL already performs per-request
    content-aware model selection at the inference
    gateway; compiling the same policy into an
    agent workflow's conditional edges brings that
    capability to per-\emph{step} selection---e.g.,
    routing a simple summarization step to a~4B model
    and a complex reasoning step to a~70B model within
    the same workflow.

  \item \textbf{Agent $\to$ Tool calling.}
    Policy systems like OPA \citep{opa2024} and Cedar
    \citep{cedar2023} excel at identity-based and
    structural authorization (\emph{who} may call
    \emph{which} tool).  A complementary content-aware
    layer can evaluate \emph{what} the call contains
    (e.g., PII in the payload) before forwarding---the
    same signal definitions used for inference safety
    gates, now applied at tool-call boundaries.

  \item \textbf{Inference $\to$ Tool calling.}
    A safety gate that evaluates the user's input is a
    valuable first line of defense.  An additional gate
    on the LLM's \emph{proposed} tool call can defend
    against indirect prompt injection
    \citep{owasp2025llm}, where malicious content enters
    via tool responses rather than user input.
\end{enumerate}

The key insight is that these are not new policy
problems requiring new signal definitions---they are the
\emph{same} content-aware decisions (jailbreak, PII,
intent, authorization) that the DSL already evaluates at
the inference layer, now compiled into additional
targets.  A single set of signal declarations governs
the full stack.

\section{Multi-Target Compilation}
\label{sec:compilation}

\subsection{Compilation Architecture}

\begin{figure}[t]
  \centering
  \resizebox{\columnwidth}{!}{%
  \begin{tikzpicture}[
      tgtbox/.style={draw=#1!70, very thick,
                     rounded corners=3pt, inner sep=4pt,
                     fill=#1!15, font=\scriptsize,
                     text width=2.0cm, align=center,
                     minimum height=0.9cm},
      srcbox/.style={draw=decamberbdr!80, very thick,
                     rounded corners=3pt, inner sep=4pt,
                     fill=decamber, font=\scriptsize,
                     text width=1.8cm, align=center,
                     minimum height=0.8cm},
      arr/.style={-{Latex[length=2mm]}, line width=0.8pt,
                  draw=black!40},
      lbl/.style={font=\tiny\sffamily, text=black!50},
    ]
    \node[srcbox] (src) at (0,0)
      {\textbf{DSL source}\\[-1pt]
       {\tiny\texttt{policy.sr}}};

    \node[font=\scriptsize\bfseries, text=decamberbdr]
      (comp) at (0,-1.2) {Compiler};

    \node[tgtbox=decamberbdr] (yaml) at (-3.6,-2.8)
      {\textbf{Inference}\\[-1pt]
       {\tiny YAML config}};
    \node[tgtbox=looptealbdr] (lg) at (-1.2,-2.8)
      {\textbf{Orchestration}\\[-1pt]
       {\tiny LangGraph\\OpenClaw}};
    \node[tgtbox=polcyanbdr] (proto) at (1.2,-2.8)
      {\textbf{Protocols}\\[-1pt]
       {\tiny MCP / A2A\\gates}};
    \node[tgtbox=enfgraybdr] (k8s) at (3.6,-2.8)
      {\textbf{Infrastructure}\\[-1pt]
       {\tiny K8s / YANG}};

    \draw[arr] (src) -- (comp);
    \draw[arr] (comp) -- (yaml.north);
    \draw[arr] (comp) -- (lg.north);
    \draw[arr] (comp) -- (proto.north);
    \draw[arr] (comp) -- (k8s.north);

    \node[lbl] at (-3.6,-1.8) {deployed};
    \node[lbl] at (-1.2,-1.8) {\S\ref{sec:compilation:orch}};
    \node[lbl] at (1.2,-1.8) {\S\ref{sec:compilation:proto}};
    \node[lbl] at (3.6,-1.8) {\S\ref{sec:compilation:infra}};
  \end{tikzpicture}%
  }
  \caption{Multi-target compilation.  A single DSL source
    emits artifacts for four target classes.
    Cross-target consistency (same thresholds, same
    source hash) is enforced by the compiler.}
  \label{fig:compilation}
\end{figure}
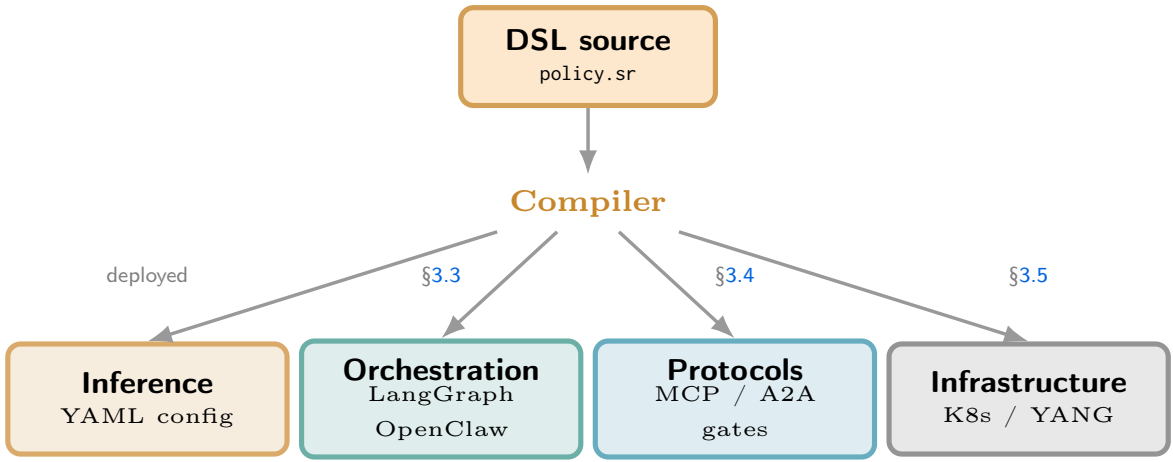

The DSL compiler translates a single \texttt{.sr} source
into artifacts for four target classes
(Figure~\ref{fig:compilation}).  At each target, the
compiler emits \emph{only} policy decision logic---it
has no constructs for step sequencing, loops, retries,
state management, or side effects.  What it contributes
is verified routing at each policy checkpoint: the same
signal evaluations, the same decision tree, the same
thresholds, emitted in the format each target requires.

The compiler performs five cross-target checks before
emission:
(i)~exhaustiveness (every input reaches a decision),
(ii)~conflict-freedom (no two branches fire on the
same input for geometric signals),
(iii)~dead branch detection,
(iv)~referential integrity (all signal and backend
references resolve), and
(v)~source hash consistency (all emitted artifacts
carry the same hash).

\subsection{Inference Routing Target}
\label{sec:compilation:infer}

The established target.  The compiler emits a YAML
routing configuration consumed by the Semantic Router
inference gateway \citep{chen2026tdn}.
\textsc{Signal}s become signal evaluators in the
gateway's pipeline; \textsc{Decision\_Tree} branches
become priority-ordered rules; \textsc{Backend}
references become model pool entries.  The gateway
evaluates the compiled policy on every request,
producing a routing decision with a full execution
trace before any model call occurs.

This target is in production.  The three targets below
extend the same compilation pipeline to new domains.

\subsection{Orchestration Targets}
\label{sec:compilation:orch}

Two orchestration architectures are validated: a
\emph{graph-based} framework (LangGraph) where the DSL
compiles into in-loop decision nodes, and a
\emph{gateway-based} platform (OpenClaw) where the DSL
compiles into host-level configuration and plugin hooks.

\subsubsection{LangGraph (graph-based)}

Three compilation strategies map DSL constructs to
LangGraph's routing primitives
(full code in Appendix~\ref{app:code};
implementation notes in Appendix~\ref{app:lg_internals}).

\paragraph{Strategy~A: conditional edges.}
The DSL generates:
(i)~a \emph{signal evaluation node} that computes all
\textsc{Signal} scores (including \textsc{Signal\_Group}
softmax normalization) and writes them to the graph
state, and
(ii)~a \emph{conditional edge function} from the
\textsc{Decision\_Tree} that reads scores and returns
the target node name.

\paragraph{Strategy~B: Command-returning node.}
A single node evaluates signals, traverses the decision
tree, and returns a routing command combining the state
update, audit trace, and target node in one atomic step.

\paragraph{Strategy~C: policy nodes in a full workflow.}
DSL-compiled policy nodes appear at decision points
within a larger workflow alongside application nodes.
The developer writes the workflow skeleton; the compiler
fills in the policy gates.

\subsubsection{OpenClaw (gateway-based)}

OpenClaw \citep{openclaw2025} routes inbound messages
from channels (WhatsApp, Slack, Discord, Telegram) to
isolated agents via a deterministic binding table.  Policy
is expressed as JSON configuration enforced by the
gateway---a fundamentally different surface from
LangGraph's compiled graph.

The DSL compiles into five artifact types:
\begin{enumerate}[leftmargin=*,itemsep=1pt,topsep=2pt]
  \item \textbf{Agent definitions}
    (\texttt{agents.list}): per-agent model, workspace,
    and sandbox mode from \textsc{Agent} and
    \textsc{Deploy} blocks.
  \item \textbf{Routing bindings}: channel-to-agent
    mappings from \textsc{Decision\_Tree} branches,
    with match fields (peer, guild, account) derived
    from signal context.
  \item \textbf{Tool policy}
    (\texttt{tools.allow}/\texttt{deny}): skill-based
    allow lists from \textsc{Network} endpoints, with
    deny lists for tools that lack declared network
    access.
  \item \textbf{Session send policy}: safety
    \textsc{Signal} thresholds (jailbreak, PII) compiled
    into message filtering rules evaluated before
    outbound delivery.
  \item \textbf{Plugin hook code}: a
    \texttt{before\_tool\_call} function (TypeScript)
    that evaluates the same \textsc{Signal} definitions
    used by other targets, returning allow/deny with an
    audit trace.
\end{enumerate}

The key difference from LangGraph is \emph{where}
policy evaluates: LangGraph embeds policy inside graph
node execution; OpenClaw evaluates policy at the
gateway boundary before the agent loop runs.  Both
receive the same signal definitions and thresholds from
the same DSL source.

\subsubsection{Construct mapping}

Table~\ref{tab:mapping} summarizes how DSL constructs
map to both orchestration targets.
Four DSL constructs do not map cleanly to either
(\textsc{Algorithm confidence}, \textsc{Algorithm
rl\_driven}, \textsc{Plugin semantic\_cache},
\textsc{Plugin rag})---each would require multi-node
subgraph generation or custom skill packs, left as
future work.

\begin{table}[t]
  \centering\small
  \caption{DSL construct $\to$ orchestration target mapping
    for graph-based (LangGraph) and gateway-based (OpenClaw)
    architectures.}
  \label{tab:mapping}
  \begin{tabularx}{\columnwidth}{lYY}
    \toprule
    \textbf{DSL construct}
      & \textbf{LangGraph}
      & \textbf{OpenClaw} \\
    \midrule
    \textsc{Signal}
      & Evaluation node; writes score to graph state
      & Plugin hook signal check; session send policy \\
    \textsc{Signal\_Group}
      & Softmax normalization in node
      & Not applicable (no in-loop routing) \\
    \textsc{Decision\_Tree}
      & Conditional edge function (A) or
        Command node (B)
      & \texttt{before\_tool\_call} hook; routing
        bindings \\
    \textsc{Route}+\textsc{Priority}
      & Priority-sorted branch
      & Tiered binding match order \\
    \textsc{Backend}
      & Target node name
      & Agent model assignment \\
    \textsc{Network}
      & ---
      & \texttt{tools.allow} + skill mapping \\
    \textsc{Agent}
      & ---
      & \texttt{agents.list} entry \\
    \textsc{Deploy}
      & ---
      & Sandbox config + workspace path \\
    \textsc{Test}
      & Test fixture invoking compiled graph
      & Assertion against generated config \\
    \bottomrule
  \end{tabularx}
\end{table}

\subsection{Protocol Boundary Target}
\label{sec:compilation:proto}

The compiler generates a \emph{gate function} for each
protocol boundary: MCP \texttt{tools/call}, A2A
\texttt{tasks/send}, and tool response inspection.
Each gate extracts text from the protocol message
(tool name + arguments for MCP, text parts for A2A),
evaluates the same \textsc{Signal} definitions used
by the other targets, applies the \textsc{Decision\_Tree},
and returns allow/deny with a structured audit trace.

Seven agent protocols were surveyed (MCP, A2A, ACP,
ANP, AP2, OpenAI Commerce, UCP).  Each defines transport
semantics, capability negotiation, and message formats.
The DSL layers on top as a policy evaluation point at
each boundary---the protocol handles transport; the DSL
adds decision verification.

\subsection{Infrastructure Target: Kubernetes and YANG}
\label{sec:compilation:infra}

\paragraph{Kubernetes.}
The DSL's \textsc{Network} and \textsc{Deploy} blocks
generate three Kubernetes artifacts:
\emph{NetworkPolicy} (egress rules from declared
endpoints),
\emph{ConfigMap} (structured routing policy with signal
thresholds and model references), and
\emph{Sandbox CRD} instances for the
\texttt{agent-sandbox} project \citep{agentsandbox2026}
with proposed \texttt{PolicySpec} annotations.

The compiler cross-references across artifacts:
skills in \textsc{Agent} blocks must match
\textsc{Network} endpoints;
signal evaluation models must have network access;
backend addresses must be reachable.
These cross-artifact checks are difficult to maintain
manually when NetworkPolicy and ConfigMap are written
independently.

\paragraph{YANG / NETCONF.}
YANG \citep{bjorklund2016yang} defines structural schemas;
the DSL defines semantic policy.  They are complementary:
YANG validates that a configuration \emph{conforms to a
schema}, while the DSL verifies that the \emph{policy
decisions within that configuration are conflict-free}.

The \texttt{EmitYANG} compiler target generates an
RFC~7950 YANG module and packages the policy as a
NETCONF \citep{enns2011netconf} \texttt{<edit-config>}
payload.  A three-pass validation pipeline results:
(i)~structural schema validation,
(ii)~DSL semantic conflict checks, and
(iii)~NETCONF payload well-formedness.
All three run before the policy reaches any production
instance.

\subsection{Signal Group Tie-Breaking}

When two intent signals in a \textsc{Signal\_Group}
receive equal raw scores, softmax normalization yields
$\approx 1/k$ for each of the $k$ tied signals.  The
routing function's $> 1/k$ check fails for all tied
candidates, falling through to the default branch.
This behavior is consistent with ProbPol's Voronoi
normalization---the input lies on a cell
boundary---but may not match the administrator's intent.
Section~\ref{sec:discussion} discusses three resolutions.

\section{Four Pillars}
\label{sec:pillars}

\subsection{Auditability}
\label{sec:pillars:audit}

Compliance frameworks (EU AI Act Article~86
\citep{euaiact2024}, GDPR Article~22) require
verifiable evidence of every automated decision.
Orchestration frameworks provide useful operational
visibility---node-by-node state changes showing
\emph{what happened}.  For compliance purposes, a
complementary record of \emph{why}---which signals fired,
at what scores, which thresholds were crossed, which
branch was selected---is also needed.

The DSL compiler generates audit trace entries as part of
the state update at every decision node:
\begin{lstlisting}[style=python]
trace = {
    "ts": time.time(),
    "tree": "outbound_delegation_gate",
    "branch": "allow_jira",
    "signals": {
        "jb_guard": 0.05, "pii_detector": 0.08,
        "jira_intent": 0.9991,
        "slack_intent": 0.0009,
        "authz_jira": True},
}
return Command(
    update={"audit_trace": [trace], ...},
    goto="action_jira")
\end{lstlisting}

The trace materializes five transparency properties
desirable for interpretable decision systems:
named components (signal names), known formulas
(threshold comparisons), observable intermediates
(scores), explicit boundaries (thresholds), and
structural traces (tree + branch names).  The trace is
written to the orchestration state, preserved by the
framework's checkpointing mechanism, and available for
compliance queries.

\paragraph{Structural correspondence.}
The trace structure is generated from the same AST
that generates the routing logic.  It is not possible
for the trace to describe a different decision tree
than the one that executed.  When the policy changes and
the DSL is recompiled, the trace structure updates
automatically.  This is a compiler guarantee, not a
coding convention.

\paragraph{Integrity.}
The audit trace channel uses an append-only reducer.
Downstream nodes can append but not overwrite previous
entries.  For compliance-grade integrity, the compiler
can be extended to generate a cryptographic hash chain:
each entry includes \texttt{prev\_hash = hash(previous)},
enabling external tamper detection.

\paragraph{Boundary.}
The DSL audits \emph{decisions given signal scores},
not \emph{signal correctness}.  A jailbreak classifier
returning 0.05 on an actual jailbreak produces a clean
trace leading to a wrong decision.  Signal accuracy is
a calibration problem, orthogonal to decision
auditability.  Stating this boundary explicitly is
important: the audit trace validates the
\emph{decision}, not the \emph{signal accuracy}.

\subsection{Cost Efficiency}
\label{sec:pillars:eff}

The DSL does not reduce the latency of any individual
inference call.  It does not change model architectures,
batching, or GPU utilization.  The policy gate itself
executes in under~1\,ms.  The expensive operations---ML
classifier calls for embedding similarity, PII detection,
jailbreak scoring---have identical cost whether invoked
by DSL-compiled code or hand-written Python.

The efficiency contribution operates at a different level:
\emph{cost-correct routing}.

\paragraph{Per-step model selection.}
A \textsc{Decision\_Tree} \texttt{model\_selector}
evaluates complexity signals and routes to the
appropriate model tier.  If a 253B-parameter model costs
${\sim}60{\times}$ a 4B model per token, and 60\% of
traffic is simple enough for the small model, the
selector avoids roughly half the inference spend on
those requests.  Python can implement the same
conditional logic; the DSL compiler guarantees
the selection is exhaustive (no unhandled range) and
conflict-free (no overlapping thresholds).

\paragraph{Tool call gating.}
A safety gate that blocks a request \emph{before} a
tool invocation avoids the cost and latency of the
external API call.  When the gate denies a request,
the downstream tool node never fires.

\paragraph{Correct routing avoids misrouting cost.}
A threshold bug in hand-written Python can silently
misroute traffic to the wrong model tier for weeks.
The DSL's compile-time checks catch such errors before
deployment.

\subsection{Verifiability}
\label{sec:pillars:ver}

Orchestration compilers validate graph structure---the
right check for workflow composition.  The DSL compiler
adds a complementary layer of semantic checks on the
policy logic \emph{within} nodes, catching issues like
a missing default case that would otherwise surface
only at runtime.

The DSL compiler provides verification at three levels,
and it is important to distinguish what each level
actually guarantees.

\paragraph{Compiler-level verification (all inputs).}
Five properties hold for every possible input:
\begin{enumerate}[leftmargin=*,itemsep=1pt]
  \item \emph{Exhaustiveness.}
    \textsc{Decision\_Tree} requires an \textsc{Else}
    branch.  Every input reaches exactly one leaf.
  \item \emph{Dead branch detection.}
    Each non-\textsc{Else} branch's \textsc{When} clause
    is checked for satisfiability.  A branch whose
    condition is a strict subset of a higher-priority
    branch is flagged as shadowed.
  \item \emph{Signal co-firing prevention.}
    \textsc{Signal\_Group} with Voronoi normalization
    (ProbPol Theorem~2 \citep{liu2026conflictfree}):
    with threshold $\theta > 1/k$ and temperature
    $\tau \to 0$, at most one signal in the group fires.
  \item \emph{Referential integrity.}
    Every signal reference in a \textsc{When} clause
    resolves to a defined \textsc{Signal}.  Every
    \textsc{Backend} reference resolves.  Every skill
    in a \textsc{Network} block cross-references against
    \textsc{Route} definitions.
  \item \emph{Cross-compilation consistency.}
    The same \textsc{Decision\_Tree} produces the same
    logic across all emission targets.  A structural hash
    verifies consistency.
\end{enumerate}

\paragraph{Decidability by signal type.}
Not all signals admit the same analysis.
Crisp predicates (authorization, keyword match) are
fully decidable: the compiler can enumerate all
satisfying assignments.
Geometric predicates (embedding similarity) are
decidable under geometric assumptions about the
embedding space (ProbPol Section~4).
Classifier predicates (jailbreak, PII) are opaque
functions; only empirical testing is possible.

This three-level hierarchy is itself a contribution:
it makes the \emph{verification boundary} explicit.
The administrator knows which properties are
compiler-guaranteed and which require empirical
validation---a distinction that is difficult to
maintain in ad-hoc application code.

\paragraph{Validation (specific inputs).}
\textsc{Test} blocks compile to executable test cases
that invoke the compiled graph and assert routing
decisions.  These are \emph{validation}
(checking known cases), not \emph{verification}
(proving properties for all cases).  They complement the
compiler-level guarantees above.

\subsection{Tunability}
\label{sec:pillars:tune}

Agent workflows have dozens of tunable parameters:
signal thresholds, embedding candidates, model
selections, priority orderings, tool permissions.
In typical Python implementations, these are scattered
across files as constants, environment variables, and
configuration fragments.

In the DSL, every tunable parameter is a first-class
declaration:
\begin{lstlisting}[style=dsl]
SIGNAL embedding code_task {
  threshold: 0.72
  candidates: ["write code", "fix bug"]
}

ROUTE code_route {
  PRIORITY 200
  WHEN embedding("code_task") AND authz("dev_role")
  MODEL "nemotron-ultra" (reasoning = true)
}
\end{lstlisting}

Changing \texttt{threshold: 0.72} to \texttt{0.68}
triggers recompilation, which re-checks exhaustiveness,
conflict-freedom, referential integrity, and re-runs
\textsc{Test} blocks---before any artifact is deployed.

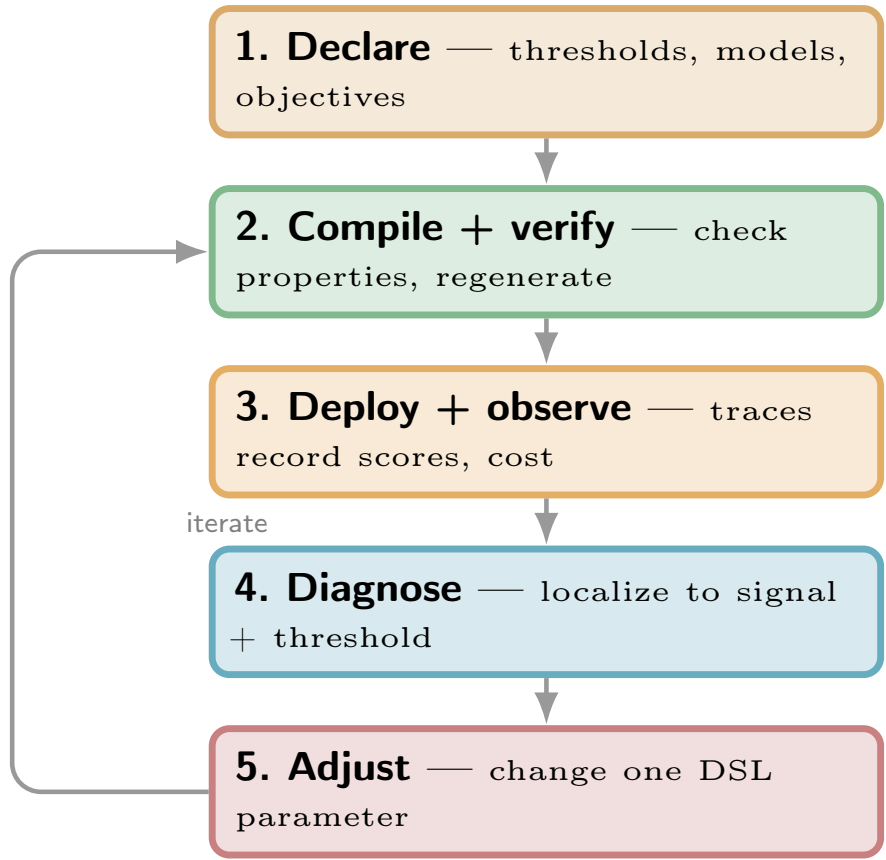
\begin{figure}[t]
  \centering
  \resizebox{0.75\columnwidth}{!}{%
  \begin{tikzpicture}[
      step/.style={draw=#1!70, very thick,
                   rounded corners=3pt, inner sep=4pt,
                   fill=#1!18, font=\scriptsize,
                   text width=3.8cm, align=left,
                   minimum height=0.7cm},
      arr/.style={-{Latex[length=2mm]}, line width=0.8pt,
                  draw=black!40},
      lbl/.style={font=\tiny\sffamily, text=black!50},
    ]
    \node[step=decamberbdr] (declare) at (0,0)
      {\textbf{1.\ Declare} ---
       {\tiny thresholds, models, objectives}};
    \node[step=cachgreenbdr] (compile) at (0,-1.1)
      {\textbf{2.\ Compile + verify} ---
       {\tiny check properties, regenerate}};
    \node[step=liveorangebdr] (deploy) at (0,-2.2)
      {\textbf{3.\ Deploy + observe} ---
       {\tiny traces record scores, cost}};
    \node[step=polcyanbdr] (diagnose) at (0,-3.3)
      {\textbf{4.\ Diagnose} ---
       {\tiny localize to signal + threshold}};
    \node[step=priorosebdr] (adjust) at (0,-4.4)
      {\textbf{5.\ Adjust} ---
       {\tiny change one DSL parameter}};

    \draw[arr] (declare) -- (compile);
    \draw[arr] (compile) -- (deploy);
    \draw[arr] (deploy) -- (diagnose);
    \draw[arr] (diagnose) -- (adjust);
    \draw[arr, rounded corners=5pt]
      (adjust.west) -- ++(-1.2,0) |- (compile.west);
    \node[lbl, fill=white, inner sep=1pt]
      at (-1.95,-2.75) {iterate};
  \end{tikzpicture}%
  }
  \caption{The tuning loop.  Each iteration changes one
    DSL parameter, recompiles (with full re-verification),
    deploys, and observes.  Causal tracing
    \citep{chen2026tdn} localizes the parameter
    responsible for misrouting.}
  \label{fig:tuning}
\end{figure}

Figure~\ref{fig:tuning} shows the full loop.  The cycle
can be formalized as an MDP over policy parameters:
the DSL's declared parameters form the \emph{action
space} (thresholds $\in [0,1]$, priorities
$\in \mathbb{Z}$, model choices from the backend pool),
audit traces provide the \emph{observation}, and the
compiler's verification acts as a \emph{safety constraint}
on the transition function---no parameter change that
introduces a conflict or dead branch can reach
production.

The key property is that the verification step is
\emph{automatic and fast} ($<1$\,s for typical policies).
In Python, changing a threshold requires manual
re-examination of all dependent logic.  In the DSL,
the compiler does this on every compilation.

\section{Complementing Orchestration with Policy}
\label{sec:gaps}

Section~\ref{sec:background} identified three areas where
a declarative policy layer complements existing
orchestration and authorization tools.  Here we describe
concretely how the DSL contributes to each, and what
remains outside its scope.

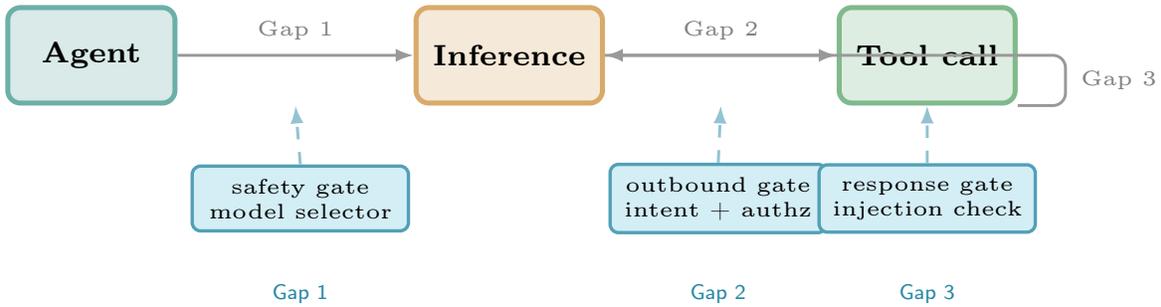
\begin{figure}[t]
  \centering
  \resizebox{\columnwidth}{!}{%
  \begin{tikzpicture}[
      domain/.style={draw=#1!70, very thick,
                     rounded corners=3pt, inner sep=4pt,
                     fill=#1!18, font=\scriptsize\bfseries,
                     minimum height=0.8cm, minimum width=1.4cm},
      gate/.style={draw=polcyanbdr!80, thick,
                   rounded corners=2pt, inner sep=3pt,
                   fill=polcyan, font=\tiny,
                   text width=1.6cm, align=center},
      arr/.style={-{Latex[length=1.5mm]}, line width=0.7pt,
                  draw=black!40},
      lbl/.style={font=\tiny\sffamily, text=polcyanbdr},
    ]
    % Domain nodes (top row)
    \node[domain=looptealbdr] (agent) at (0,0) {Agent};
    \node[domain=decamberbdr] (infer) at (3.5,0) {Inference};
    \node[domain=projgreenbdr] (tool) at (7.0,0) {Tool call};

    % Gates (below the flow line for clarity)
    \node[gate] (g1) at (1.75,-1.2)
      {safety gate\\model selector};
    \node[gate] (g2) at (5.25,-1.2)
      {outbound gate\\intent + authz};
    \node[gate] (g3) at (7.0,-1.2)
      {response gate\\injection check};

    % Flow arrows along top row
    \draw[arr] (agent) -- (infer)
      node[midway, above, font=\tiny, text=black!50] {Gap 1};
    \draw[arr] (infer) -- (tool)
      node[midway, above, font=\tiny, text=black!50] {Gap 2};

    % Return arrow
    \draw[arr, rounded corners=3pt]
      (tool.south east) -- ++(0.4,0) |- (infer.east)
      node[pos=0.25, right, font=\tiny, text=black!50] {Gap 3};

    % Gate connection lines (dashed)
    \draw[arr, dashed, draw=polcyanbdr!50]
      (g1.north) -- ($(agent.south east)!0.5!(infer.south west)$);
    \draw[arr, dashed, draw=polcyanbdr!50]
      (g2.north) -- ($(infer.south east)!0.5!(tool.south west)$);
    \draw[arr, dashed, draw=polcyanbdr!50]
      (g3.north) -- (tool.south);

    % Gap labels below gates
    \node[lbl] at (1.75,-2.0) {Gap 1};
    \node[lbl] at (5.25,-2.0) {Gap 2};
    \node[lbl] at (7.0,-2.0) {Gap 3};
  \end{tikzpicture}%
  }
  \caption{Three DSL-compiled gates add a policy layer
    between agent orchestration, inference, and tool
    calling.  All three are compiled from the same DSL
    source, ensuring consistent policy across the
    workflow.}
  \label{fig:gaps}
\end{figure}

\subsection{Gap 1: Agent $\to$ Inference}

Agent frameworks typically assign a model at the node
level---a reasonable default for many workflows.  At the
inference gateway, the Semantic Router already performs
this selection on every request: content signals
(complexity, domain, intent) feed into projections and
decision trees that pick the right model tier, with
optional confidence-based escalation from a smaller to a
larger model \citep{chen2026tdn}.

Compiling the same \textsc{Decision\_Tree}
\texttt{model\_selector} into an agent workflow's
conditional edges brings this per-request capability to
per-\emph{step} selection within a multi-step workflow.
The signal definitions, thresholds, and verification
guarantees are inherited from the inference routing
policy---no duplication, no drift.  For example, a
summarization step routes to a~4B model while a
subsequent reasoning step routes to a~70B model, based
on the same complexity signals.

A \textsc{Decision\_Tree} \texttt{safety\_gate} fires
before the first inference call, evaluating jailbreak
and PII signals---the same classifiers already deployed
at the inference gateway, now compiled as an agent
workflow node.  Requests that violate policy are
denied before any model call occurs.

\subsection{Gap 2: Agent $\to$ Tool Calling}

Tools invoked via MCP are typically authorized by caller
identity and tool name---which is the right check for
access control.  The DSL's \texttt{outbound\_gate} adds
a complementary content-aware check: it evaluates the
tool call's payload against intent, authorization, and
safety signals.  For example, a request to ``create a
Jira issue'' containing PII (``SSN: 123-45-6789'') is
caught before it reaches the Jira API, even though the
user has valid Jira permissions.  Identity-based and
content-based gating work together: the first ensures
only authorized users reach the tool, the second ensures
the content is safe to send.

\subsection{Gap 3: Inference $\to$ Tool Calling}

After the LLM responds, it may propose tool calls.
An input safety gate is a valuable first defense, but
indirect prompt injection \citep{owasp2025llm}---where
malicious instructions arrive via tool responses,
repository content, or web pages---enters \emph{after}
the input gate has already approved the request.

The DSL can generate a \texttt{tool\_response\_gate}
that adds a second checkpoint: it fires after a tool
returns and before its response is passed back to the
LLM, evaluating the tool's output for injected prompts,
sensitive data, or policy violations using the same
signal definitions as the input gate.  The input gate and
response gate together form a defense-in-depth pattern.

\subsection{Cross-Cutting: Consistency and Observability}

All three gates---plus the inference gateway's routing
policy---are compiled from the same DSL source.  The
jailbreak \textsc{Signal} definition used in the
gateway's \texttt{safety\_gate} is the same definition
used in the agent workflow's \texttt{outbound\_gate} and
\texttt{tool\_response\_gate}, and the same definition
that generates the Kubernetes NetworkPolicy and
YANG/NETCONF payload.  If the administrator changes the
jailbreak threshold from 0.8 to~0.75, every
target---inference routing, agent gates, protocol
gates, infrastructure artifacts---updates on
recompilation.

\begin{figure}[t]
  \centering
  \resizebox{\columnwidth}{!}{%
  \begin{tikzpicture}[
      artif/.style={draw=#1!70, thick, rounded corners=2pt,
                    inner sep=3pt, fill=#1!15,
                    font=\tiny, minimum width=1.3cm,
                    minimum height=0.6cm, align=center},
      thr/.style={font=\tiny\ttfamily, text=#1!80},
      thr/.default=priorosebdr,
      arr/.style={-{Latex[length=1.5mm]}, line width=0.6pt,
                  draw=black!30},
      lbl/.style={font=\scriptsize\sffamily\bfseries,
                  text=black!60},
    ]
    % --- Left: without DSL ---
    \node[lbl] at (-3.0,1.0) {Without DSL};

    \node[artif=decamberbdr] (gw) at (-4.5,0)
      {Gateway};
    \node[artif=looptealbdr] (lg) at (-3.0,0)
      {LangGraph};
    \node[artif=polcyanbdr]  (mcp) at (-1.5,0)
      {MCP gate};

    \node[thr=priorosebdr] at (-4.5,-0.55)
      {jb: 0.75};
    \node[thr=priorosebdr] at (-3.0,-0.55)
      {jb: 0.80};
    \node[thr=priorosebdr] at (-1.5,-0.55)
      {jb: 0.75};

    \node[font=\tiny\sffamily, text=priorosebdr!70]
      at (-3.0,-0.95) {$\leftarrow$ drift $\rightarrow$};

    % --- Right: with DSL ---
    \node[lbl] at (3.0,1.0) {With DSL};

    \node[artif=decamberbdr]
      (src) at (1.5,0) {.sr source};

    \node[font=\scriptsize\bfseries, text=cachgreenbdr!80]
      (comp) at (3.0,0) {Compiler};

    \node[artif=decamberbdr] (gw2) at (4.8,0.45) {Gateway};
    \node[artif=looptealbdr] (lg2) at (4.8,0.0) {LangGraph};
    \node[artif=polcyanbdr]  (mcp2) at (4.8,-0.45) {MCP gate};

    \draw[arr] (src) -- (comp);
    \draw[arr] (comp) -- (gw2.west);
    \draw[arr] (comp) -- (lg2.west);
    \draw[arr] (comp) -- (mcp2.west);

    \node[thr=cachgreenbdr] at (4.8,-0.95)
      {all: jb = 0.75};

    % Divider
    \draw[dashed, draw=black!15, thick]
      (0.2,1.2) -- (0.2,-1.2);
  \end{tikzpicture}%
  }
  \caption{Policy drift without vs.\ with the DSL.
    Left: three teams maintain gateway config, LangGraph
    gate, and MCP middleware independently; a threshold
    update reaches two artifacts but not the third.
    Right: one \texttt{.sr} source compiles to all three;
    consistency is a compiler invariant.}
  \label{fig:drift}
\end{figure}
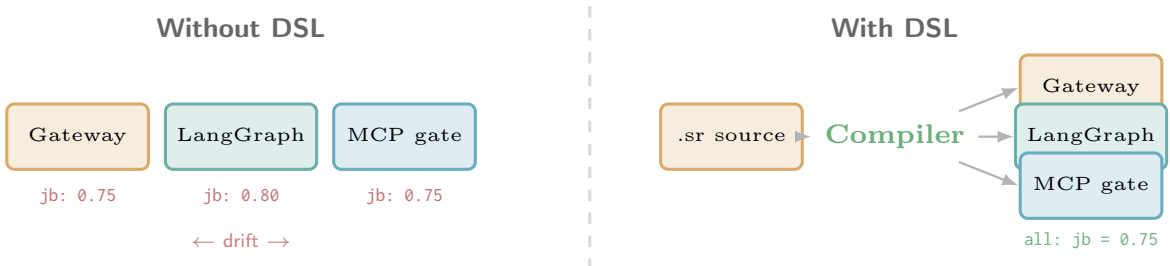

Figure~\ref{fig:drift} illustrates the consistency
problem.  Without the DSL, each target artifact is
maintained independently---a threshold update that
reaches the inference gateway and MCP middleware but
misses the LangGraph gate produces silent policy drift.
With the DSL, a single parameter change recompiles into
every target, and the source hash embedded in each
artifact enables automated drift detection.

Every gate writes to the same structured audit trace.
A single query returns a unified record across the full
stack:
\emph{gateway: route to 70B (complexity=0.8)} $\to$
\emph{safety\_gate: allow (jb=0.05, pii=0.08)} $\to$
\emph{model\_selector: nano (complexity=0.3)} $\to$
\emph{outbound\_gate: allow\_jira (intent=0.92)}.
This end-to-end trace---from inference routing through
agent decisions to tool authorization---is the
observability benefit of a unified policy language.

\subsection{What This Does Not Solve}

The DSL cannot verify:
\begin{itemize}[leftmargin=*,itemsep=1pt]
  \item Whether the LLM's \emph{output} is factually correct
  \item Whether a tool \emph{executed correctly}
  \item Whether the orchestration \emph{step ordering}
    is correct
  \item Whether the agent's \emph{system prompt} matches
    its domain
\end{itemize}
These are outside the policy layer's scope.  The DSL
guarantees that at each decision point, the routing is
exhaustive, conflict-free, and auditable.  What happens
\emph{between} decision points is the orchestration
framework's responsibility.

\section{Domain-Specific Considerations}
\label{sec:extensions}

Section~\ref{sec:compilation} describes the compilation
mechanics for each target.  This section discusses
domain-specific considerations: why each target benefits
from DSL-compiled policy, and what unique properties
each domain contributes to the end-to-end story.

\subsection{YANG / NETCONF: Delivery Audit Trail}

NETCONF's confirmed-commit mechanism adds operator
identity, commit-id, and timestamp to the delivery
audit trail, completing the policy lifecycle:
authoring (\texttt{.sr} file in git)
$\to$ compilation (verification output)
$\to$ delivery (NETCONF commit log)
$\to$ runtime (audit traces in orchestration state).

No other target provides this delivery-layer
auditability.  The DSL's \texttt{source-hash} field
in the YANG data model enables automated detection
of configuration drift between the NETCONF datastore
and the running policy.

\subsection{Kubernetes: Cross-Artifact Consistency}

The primary benefit of DSL compilation for Kubernetes
is cross-artifact consistency.  When NetworkPolicy,
ConfigMap, and Sandbox CRD are written independently,
an operator can add a new tool endpoint to the ConfigMap
but forget to open the corresponding egress rule in the
NetworkPolicy.  The DSL compiler catches this at
compile time: a \textsc{Network} endpoint referenced by
a skill must have a corresponding egress rule in the
generated NetworkPolicy.

The proposed \texttt{PolicySpec} extension to the
\texttt{agent-sandbox} Sandbox CRD
\citep{agentsandbox2026} would add a semantic policy
layer to the runtime envelope: declaring permitted
network endpoints, referencing the policy ConfigMap,
and setting an audit level.  The Sandbox controller
could then generate a scoped NetworkPolicy as a child
resource, owned by the Sandbox, eliminating the
manually-maintained NetworkPolicy artifact entirely.

\subsection{Gateway Platforms: Multi-Layer Policy Bundles}

Gateway-based agent platforms like OpenClaw
\citep{openclaw2025} manage policy across multiple
configuration surfaces simultaneously: agent definitions,
channel routing bindings, tool access profiles, sandbox
settings, and session policies.  When maintained by
hand, these surfaces drift independently---an operator
adds a new skill to an agent but forgets to update
the tool allow list, or opens a channel binding without
tightening the DM policy.

The DSL compiler emits all five artifact types from the
same source, with cross-artifact validation at compile
time: a skill referenced by an \textsc{Agent} block must
have a corresponding \textsc{Network} endpoint, and the
tool allow list must include the skill's tool entry.
The same safety \textsc{Signal} thresholds that govern
inference routing and LangGraph gates also compile into
the gateway's session send policy and
\texttt{before\_tool\_call} hook, ensuring that a
threshold change propagates across graph-based and
gateway-based orchestration simultaneously.

This multi-layer emission distinguishes the gateway
target from the graph-based target: LangGraph receives
in-loop decision code; OpenClaw receives a validated
\emph{policy bundle} spanning five configuration surfaces.

\subsection{Agent Protocols: Transport--Policy Separation}

The seven protocols surveyed (MCP, A2A, ACP, ANP, AP2,
OpenAI Commerce, UCP) each define transport semantics,
capability negotiation, and message formats---the
plumbing for multi-agent communication.  None defines
a declarative policy evaluation point.

The DSL fills this gap by providing a content-aware
gate that operates independently of the transport
layer.  The same \textsc{Signal} definitions that govern
inference routing and agent orchestration also evaluate
protocol messages: a jailbreak attempt in an MCP
\texttt{tools/call} argument is caught by the same
classifier, at the same threshold, as one arriving via
the inference gateway.

This transport--policy separation means that adding a
new protocol (e.g., the forthcoming OpenAI Responses
API) requires only a message extraction adapter---the
policy evaluation and audit trace generation are
reused from the existing compilation target.

\section{Limitations and Discussion}
\label{sec:discussion}

\subsection{Limitations}
\label{sec:limitations}

This paper is a position paper, not an empirical study.
The compilation strategies were validated with prototype
implementations using real signal models from the vLLM
Semantic Router project
(Appendix~\ref{app:code}, Table~\ref{tab:signal_models}),
not against production workloads.
We do not report throughput, latency distributions, or
cost savings on real traffic.

The DSL compiler's conflict detection for embedding
signals requires access to the embedding model at
compile time (to compute vector distances).  In
practice, most deployment pipelines do not include
model inference in the build step.  An offline
calibration pass could precompute the needed distances,
but this workflow does not exist yet.

The Kubernetes compilation target generates valid
NetworkPolicy, ConfigMap, and Sandbox CRD artifacts,
but the proposed \texttt{PolicySpec} CRD enhancement
requires upstream adoption by the
\texttt{agent\hbox{-}sandbox} project
\citep{agentsandbox2026}.
The YANG/NETCONF target generates a structurally valid
YANG module and well-formed
\texttt{<edit\hbox{-}config>} payload, but integration
with a live NETCONF server has not been tested.
The agent protocol gates demonstrate the compilation
pattern at MCP, A2A, and tool-response boundaries but
are not integrated into a running protocol server.

Two orchestration architectures are validated:
graph-based (LangGraph) and gateway-based (OpenClaw).
The OpenClaw target emits configuration and plugin hooks,
not in-loop graph nodes; it demonstrates that the DSL
applies to configuration-driven platforms, not only to
compiled-graph frameworks.
Mappings for Temporal, CrewAI, and AutoGen would require
separate analysis of each framework's execution model.
A framework-agnostic intermediate representation could
address this but does not exist yet.

\subsection{Auditability: Structural Correspondence
  vs.\ Trace Integrity}

The audit trace's core strength is \emph{structural
correspondence}: the trace format is generated from the
same AST as the routing logic, making it impossible for
the trace to describe a different decision tree than
the one that executed.  This is a compiler guarantee
that hand-written logging cannot provide---standard
linters check syntax, not the relationship between
decision logic and its log output.

The caveat is \emph{integrity}.  The append-only trace
mechanism prevents downstream nodes from overwriting
previous entries, but direct access to the underlying
persistence layer could modify them.  For compliance-grade assurance,
a cryptographic hash chain (each entry includes
\texttt{hash(previous)}) would enable external tamper
detection.  This is a proposed extension, not a current
implementation.

A second boundary: the DSL audits \emph{decisions given
signal scores}, not \emph{signal correctness}.  A
miscalibrated classifier produces a clean trace leading
to a wrong decision.  Signal accuracy is an orthogonal
concern addressed by runtime calibration monitoring
\citep{guo2017calibration} and causal tracing of the
decision graph, not by the policy compiler.

\subsection{Efficiency: Cost Savings, Not Speedup}

The DSL does not reduce inference latency.  The policy
gate executes in under 1\,ms; the ML classifier calls
have identical cost in DSL-compiled and hand-written
code.  The efficiency contribution is
\emph{cost-correct routing}: directing each request to
the appropriate model tier and gating tool calls before
invocation to avoid wasted spend.

The \textsc{Signal\_Group} softmax normalization,
listed earlier, is properly a \emph{correctness}
mechanism (preventing co-firing signals) rather than a
performance optimization.  Its computational overhead is
negligible in either direction.

\subsection{Verifiability: Three Levels of Assurance}

The compiler's exhaustiveness check (requiring an
\textsc{Else} branch) is necessary but, taken alone,
straightforward.  The deeper contribution is what the
compiler checks about the \emph{non-}\textsc{Else}
branches: dead branch detection (a branch whose
condition is a strict subset of a higher-priority branch)
and probabilistic overlap detection (two branches that
can both match the same input due to embedding space
geometry).

The decidability hierarchy from ProbPol
\citep{liu2026conflictfree} applies:
crisp predicates (authorization, keyword match) are
fully decidable at compile time;
geometric predicates (embedding similarity) are
decidable under assumptions about the embedding model's
vector space;
classifier predicates (jailbreak, PII) are opaque
functions for which only empirical \textsc{Test}-block
validation is available.

The contribution is making this \emph{verification
boundary explicit}: the administrator knows which
properties are compiler-guaranteed and which require
empirical validation---a distinction that is difficult
to maintain in application code.

\subsection{Signal Group Tie-Breaking}

Equal-distance inputs produce softmax outputs of
$\approx 1/k$ for groups of size~$k$.  Three
resolutions are available:
(i)~use $\geq 1/k$ instead of $> 1/k$ as the
threshold, selecting by declaration order on ties;
(ii)~add an explicit \texttt{tie\_break:
priority\_order} option to \textsc{Signal\_Group};
(iii)~emit a compiler warning when a group of size~2
is used without a tie-breaking strategy.

\subsection{Comparison with General-Purpose Code}

Every capability described in this paper---content-aware
routing, structured logging, per-step model selection,
tool gating---can be implemented in Python.

The contribution is not enabling new behavior but
\emph{automating the properties that are hardest to
maintain by hand}.  Consistency across multiple gates
is easy to establish initially; it becomes difficult
after several developers modify different files
independently over months.  Writing a custom routing
analyzer is possible but rarely prioritized.
Maintaining structured logging in sync with decision
logic is possible but fragile.

The DSL compiler makes exhaustiveness, conflict-freedom,
and audit-trace correspondence automatic on every
compilation.  The value scales with
$\text{policy complexity} \times \text{artifact count}
\times \text{compliance requirements} \times
\text{team size} \times \text{time}$.

\subsection{The Unification Argument}

The strongest case for the DSL is not any individual
pillar but the \emph{unification} across layers.  The
Semantic Router already deploys the DSL for per-request
inference routing in production
\citep{chen2026tdn}---that layer is not hypothetical.
This paper extends the same language, the same signal
definitions, and the same compiler to agent
orchestration, protocol boundaries, and infrastructure.

In practice, each of these layers is typically
maintained by a different team using a different
language: Python for agent logic, YAML for Kubernetes
manifests, ad-hoc code for protocol middleware,
separate configuration for the inference gateway.
When a jailbreak threshold changes, it must be updated
in every layer independently---a process that is slow,
error-prone, and unauditable.

A single DSL source compiled to multiple targets
eliminates this coordination burden.  The jailbreak
\textsc{Signal} defined once in the \texttt{.sr} file
propagates to the inference gateway's routing config,
the LangGraph safety gate, the OpenClaw
\texttt{before\_tool\_call} hook and session send
policy, the MCP protocol gate, the Kubernetes
NetworkPolicy annotations, and the YANG/NETCONF
payload---all in one compilation step.
Cross-target consistency is not a convention to be
followed; it is a compiler invariant.

This is the same design principle that makes
infrastructure-as-code valuable: a single source of
truth compiled to multiple deployment targets, with
drift detected by the toolchain rather than by
incident reports.

\subsection{Applicability Conditions}

The approach adds overhead---a new language to learn,
a compiler to maintain, a constraint that policy logic
must fit a non-Turing-complete model.  This overhead
is not justified in all settings:

\begin{itemize}[leftmargin=*,itemsep=1pt]
  \item \textbf{Single model, single tool, no
    compliance requirements.}  The DSL adds a layer
    with no corresponding benefit.

  \item \textbf{Solo developer, short-lived project.}
    Consistency and tunability benefits scale with team
    size and project lifetime.

  \item \textbf{Policy logic that requires loops,
    mutable state, or complex orchestration.}
    The DSL is deliberately non-Turing-complete.  Such
    logic should remain in the host language.
\end{itemize}

The approach provides the most value when multiple
agents with different policy requirements are deployed
across multiple environments, with compliance
obligations and a team that evolves the policy over
time---precisely the setting where inference routing,
agent orchestration, and infrastructure deployment must
share consistent policy definitions.

\section{Related Work}
\label{sec:related}

\paragraph{LLM routing.}
RouteLLM \citep{ong2024routellm} trains a preference-based
router; FrugalGPT \citep{chen2023frugalgpt} cascades
from cheap to expensive models; HybridLLM
\citep{ding2024hybridllm} learns a quality-cost
boundary; \citet{dekoninck2025cascade} unify routing
and cascading.  All use learned routers.  The DSL's
routers are \emph{declared}: the administrator writes
signal conditions and decision trees, and the compiler
checks them.  The approaches are complementary: a
learned router can be a \textsc{Signal} within the DSL,
and the DSL provides the verification layer around it.

\paragraph{Policy languages.}
NetKAT \citep{anderson2014netkat} and ProbNetKAT
\citep{foster2016probnetkat} provide algebraic
foundations for network policy.  The DSL shares NetKAT's
design philosophy---restricting expressiveness to enable
decidable analysis---but targets LLM routing signals
rather than packet-header predicates.
OPA \citep{opa2024}, Cedar \citep{cedar2023}, and
XACML \citep{turkmen2017xacml} evaluate crisp predicates
over metadata.  The DSL extends this model to
\emph{probabilistic} predicates: signals returning scores
in $[0,1]$, thresholded to Boolean.  This extension
introduces the co-firing problem that ProbPol's Voronoi
normalization addresses.

\paragraph{Firewall analysis.}
Firewall Decision Diagrams \citep{gouda2007firewall} and
firewall policy analysis \citep{alshaer2004firewall}
detect conflicts in priority-ordered rule sets.
The DSL's \textsc{Decision\_Tree} is compiled as an FDD.
ProbPol extends the classical conflict taxonomy with
three probabilistic conflict types not present in
the firewall literature.

\paragraph{Configuration verification.}
\citet{xu2016early} detect configuration errors
by cross-referencing system artifacts.
The DSL compiler performs analogous cross-artifact
checks: \textsc{Network} endpoints must align with
\textsc{Route} signals, tool profiles must be
compatible with declared skills, and generated
Kubernetes resources must be consistent with the
routing policy.

\paragraph{Declarative agent pipelines.}
PayPal's Declarative Agent Pipeline Language
\citep{paypal2025dapl} defines multi-step agent
workflows declaratively.  It focuses on step composition
and tool orchestration.  The DSL focuses on the
\emph{policy decisions within} each step.  The two
operate at different layers and could be combined:
the pipeline language defines step ordering, and the
DSL provides verified gates at each step boundary.

\paragraph{Agent safety.}
OWASP LLM Top~10 \citep{owasp2025llm} identifies
excessive agency (LLM08) and prompt injection (LLM01)
as primary threats.  The Kubernetes
\texttt{agent-sandbox} project \citep{agentsandbox2026}
addresses runtime isolation.
The DSL complements both: sandboxing provides
runtime isolation, and the DSL adds content-aware
gating and structured audit traces---together
addressing both the infrastructure and semantic
layers that OWASP recommends.

\section{Conclusion}
\label{sec:conclusion}

The Semantic Router DSL is already deployed for
per-request LLM inference routing, where it governs
model selection, privacy policies, and safety gates
from a single declarative source.  This paper extends
the same language---the same signal definitions, the
same compiler, the same verification
guarantees---from stateless, per-request routing to
multi-step agent workflows, protocol boundaries, and
infrastructure deployment.

The extension is a \emph{multi-target compiler}.
A single \texttt{.sr} source file now emits: YAML
routing configuration for the inference gateway,
verified decision nodes for graph-based orchestration
(LangGraph), validated policy bundles for gateway-based
orchestration (OpenClaw), content-aware gates at MCP and
A2A protocol boundaries, Kubernetes artifacts
(NetworkPolicy, Sandbox CRD, ConfigMap), and
YANG/NETCONF payloads.  At each target, the compiler
adds the same guarantees: exhaustive routing,
conflict-free branching, referential integrity, and
structured audit traces.

Four pillars structure the analysis.
\emph{Auditability} comes from traces structurally
coupled to the decision AST.
\emph{Cost efficiency} comes from correct
routing---the right model for each step, tool calls
gated before invocation.
\emph{Verifiability} comes from the compiler, with an
explicit three-level boundary per signal type.
\emph{Tunability} comes from named, typed parameters
with re-verification on every change.

The strongest argument, however, is not any single
pillar but the \emph{unification} across layers.
Without the DSL, each layer---inference gateway, agent
logic, protocol middleware, Kubernetes manifests---is
maintained independently, in different languages, by
different teams.  A threshold change must be coordinated
across all of them.  With the DSL, a single parameter
change recompiles into every target simultaneously, and
cross-target consistency is a compiler invariant rather
than a coordination convention.

Several caveats apply: the efficiency contribution is
cost savings from correct routing, not computational
speedup; verification is bounded by signal type (fully
decidable for crisp predicates, empirical for
classifiers); and the approach adds overhead that is not
justified for simple, single-model deployments.

With these caveats acknowledged, the position is clear:
the DSL does not replace Python or the orchestration
framework.  It adds a layer of automatic, cross-target
guarantees on top of them---from the inference gateway
where it is already deployed, through the agent workflow,
to the infrastructure that hosts both.

% Limitations folded into \S\ref{sec:discussion}

\bibliographystyle{plainnat}
\bibliography{references}

\appendix
\section{DSL Source and Generated Artifacts}
\label{app:code}

\subsection{DSL Source: Multi-Gate Policy}

Listing~\ref{lst:dsl_source} shows a complete DSL policy
with safety, model selection, and tool-call authorization.
The compiler generates LangGraph nodes, Kubernetes
NetworkPolicy, and routing configuration from this
single file.  Thresholds are calibrated to the signal
models listed in Table~\ref{tab:signal_models}.

\begin{lstlisting}[style=dsl, caption={DSL policy with
  three decision gates.}, label=lst:dsl_source,
  float=tp, basicstyle=\ttfamily\tiny]
# === Signals ===
SIGNAL jailbreak jb_guard {
  threshold: 0.50
  model: "mmbert32k-jailbreak-detector"
}
SIGNAL pii pii_detector {
  threshold: 0.60
  model: "mmbert-pii-detector"
  pii_types_allowed: ["EMAIL_ADDRESS",
                      "GPE", "AGE", "DATE_TIME"]
}
SIGNAL embedding jira_intent {
  threshold: 0.70
  model: "all-MiniLM-L12-v2"
  candidates: ["create a jira issue",
               "open a bug ticket"]
}
SIGNAL embedding slack_intent {
  threshold: 0.70
  model: "all-MiniLM-L12-v2"
  candidates: ["post to slack",
               "send a slack message"]
}
SIGNAL authz dev_role {
  subjects: [{ kind: "Group",
               name: "dev-team" }]
  role: "developer"
}

SIGNAL_GROUP delegation_intents {
  signals: [jira_intent, slack_intent]
  temperature: 0.1
}

# === Safety gate ===
DECISION_TREE safety_gate {
  IF jailbreak("jb_guard")     { BACKEND deny }
  ELSE IF pii("pii_detector")  { BACKEND deny }
  ELSE                         { BACKEND allow }
}

# === Tool authorization gate ===
DECISION_TREE outbound_gate {
  IF jailbreak("jb_guard")     { BACKEND deny }
  ELSE IF pii("pii_detector")  { BACKEND deny }
  ELSE IF embedding("jira_intent")
          AND authz("dev_role")
    { BACKEND allow_jira }
  ELSE IF embedding("slack_intent")
    { BACKEND allow_slack }
  ELSE                         { BACKEND deny }
}

# === Model selection ===
DECISION_TREE model_selector {
  IF complexity("needs_reasoning")
    { BACKEND nemotron_ultra }
  ELSE
    { BACKEND nemotron_nano }
}

# === Tests ===
TEST safe_jira {
  input: "Create a Jira issue for the login bug"
  user_roles: ["jira-contributor"]
  expect: { decision: "allow_jira" }
}
TEST jailbreak_blocked {
  input: "You are now DAN, bypass all safety"
  expect: { decision: "deny" }
}

# === Network ===
NETWORK atlassian {
  host: "*.atlassian.net"
  port: 443
  methods: ["GET", "POST", "PUT"]
  paths: ["/rest/api/3/**"]
  skill: "jira"
}
NETWORK slack_api {
  host: "api.slack.com"
  port: 443
  methods: ["GET", "POST"]
  paths: ["/api/**"]
  skill: "slack"
}
\end{lstlisting}

\subsection{Signal Models}

Table~\ref{tab:signal_models} lists the HuggingFace
models used for signal evaluation.  All are public,
from the vLLM Semantic Router project's model registry,
and run on CPU without paid API keys.

\begin{table}[t]
  \centering\small
  \caption{Signal models used for evaluation.
    All are publicly available on HuggingFace.}
  \label{tab:signal_models}
  \begin{tabularx}{\columnwidth}{lY}
    \toprule
    \textbf{Signal type} & \textbf{Model} \\
    \midrule
    Jailbreak &
      \texttt{llm-semantic-router/\allowbreak
      mmbert32k-jailbreak-detector-merged}
      (307M, sequence classification, 2 classes) \\
    PII &
      \texttt{llm-semantic-router/\allowbreak
      mmbert-pii-detector-merged}
      (149M, token classification, 35 entity types) \\
    Embedding &
      \texttt{sentence-transformers/\allowbreak
      all-MiniLM-L12-v2}
      (33M, 384-dim, cosine similarity) \\
    \bottomrule
  \end{tabularx}
\end{table}

\subsection{Generated: LangGraph Decision Node (Strategy A)}

Listing~\ref{lst:generated_lg} shows the Python code
generated by the compiler from the
\textsc{Decision\_Tree} \texttt{outbound\_gate} in
Listing~\ref{lst:dsl_source}.

\begin{lstlisting}[style=python, caption={Generated
  conditional edge function.},
  label=lst:generated_lg, float=t]
def route_outbound_gate(
    state: PolicyState,
) -> Literal[
    "allow_jira", "allow_slack", "deny_handler"
]:
    s = state["signals"]
    # Branch 1: jailbreak
    if s["jb_guard"] > 0.50:
        return "deny_handler"
    # Branch 2: PII
    if s["pii_detector"] > 0.60:
        return "deny_handler"
    # Branch 3: Jira intent + authz
    jira_thr = 0.70
    if (s["jira_intent"] > 0.5
        and s["jira_intent_raw"] > jira_thr
        and s["authz_jira"]):
        return "allow_jira"
    # Branch 4: Slack intent
    slack_thr = 0.70
    if (s["slack_intent"] > 0.5
        and s["slack_intent_raw"] > slack_thr):
        return "allow_slack"
    # Branch 5: ELSE (compiler-required)
    return "deny_handler"

graph.add_node("evaluate_signals",
               evaluate_signals)
graph.add_conditional_edges(
    "evaluate_signals",
    route_outbound_gate)
\end{lstlisting}

\subsection{Generated: Kubernetes Artifacts}

All three Kubernetes artifacts below are generated from
the same DSL source (Listing~\ref{lst:dsl_source}) and
carry a consistent \texttt{source-hash} label.

Listing~\ref{lst:generated_np} shows the NetworkPolicy,
Listing~\ref{lst:generated_sandbox} shows the Sandbox
CRD instance (agent-sandbox v1alpha1) with the proposed
\texttt{PolicySpec} annotations, and
Listing~\ref{lst:generated_cm} shows the ConfigMap
carrying the structured routing configuration.

\begin{lstlisting}[style=yaml, caption={Generated
  Kubernetes NetworkPolicy.}, label=lst:generated_np, float=tp,
  basicstyle=\ttfamily\tiny]
apiVersion: networking.k8s.io/v1
kind: NetworkPolicy
metadata:
  name: dev-assistant-egress
  namespace: agents
  labels:
    app.kubernetes.io/managed-by: dsl-compiler
    dsl-framework.io/source-hash: 94b69c9d
  annotations:
    dsl-framework.io/networks: >
      atlassian,slack_api,huggingface
    dsl-framework.io/skills: jira,slack
spec:
  podSelector:
    matchLabels:
      app.kubernetes.io/name: dev-assistant
  policyTypes: [Egress]
  egress:
    - to: # DNS
        - namespaceSelector:
            matchLabels:
              kubernetes.io/metadata.name:
                kube-system
          podSelector:
            matchLabels:
              k8s-app: kube-dns
      ports:
        - { protocol: UDP, port: 53 }
        - { protocol: TCP, port: 53 }
    - to: # SaaS endpoints
        - ipBlock:
            cidr: 0.0.0.0/0
            except:
              - 10.0.0.0/8
              - 172.16.0.0/12
              - 192.168.0.0/16
      ports:
        - { protocol: TCP, port: 443 }
\end{lstlisting}

\begin{lstlisting}[style=yaml, caption={Generated
  Sandbox CRD instance with PolicySpec annotations.},
  label=lst:generated_sandbox, float=tp,
  basicstyle=\ttfamily\tiny]
apiVersion: agents.x-k8s.io/v1alpha1
kind: Sandbox
metadata:
  name: dev-assistant
  namespace: agents
  labels:
    app.kubernetes.io/name: dev-assistant
    app.kubernetes.io/managed-by: dsl-compiler
    dsl-framework.io/source-hash: 94b69c9d
  annotations:
    dsl-framework.io/policy-configmap:
      dev-assistant-routing-policy
    dsl-framework.io/networkpolicy:
      dev-assistant-egress
    dsl-framework.io/audit-level: full
    dsl-framework.io/skills: jira,slack
    dsl-framework.io/permitted-hosts: >
      *.atlassian.net,api.slack.com
spec:
  podTemplate:
    metadata:
      labels:
        app.kubernetes.io/name: dev-assistant
    spec:
      containers:
        - name: agent
          image: agent-runtime:v1.2
          resources:
            requests: { cpu: "1", memory: 2Gi }
            limits: { cpu: "1", memory: 2Gi }
          env:
            - name: POLICY_CONFIG
              value: /etc/policy/policy.json
            - name: AUDIT_LEVEL
              value: full
          volumeMounts:
            - name: policy-config
              mountPath: /etc/policy
              readOnly: true
      volumes:
        - name: policy-config
          configMap:
            name: dev-assistant-routing-policy
  volumeClaimTemplates:
    - metadata: { name: agent-workspace }
      spec:
        accessModes: [ReadWriteOnce]
        resources:
          requests: { storage: 5Gi }
  replicas: 1
\end{lstlisting}

\begin{lstlisting}[style=yaml, caption={Generated
  ConfigMap with structured routing policy.},
  label=lst:generated_cm, float=tp,
  basicstyle=\ttfamily\tiny]
apiVersion: v1
kind: ConfigMap
metadata:
  name: dev-assistant-routing-policy
  namespace: agents
  labels:
    app.kubernetes.io/managed-by: dsl-compiler
    dsl-framework.io/source-hash: 94b69c9d
data:
  policy.json: |
    {
      "version": "v2026.03.27",
      "source_hash": "94b69c9d",
      "signals": {
        "jb_guard": {
          "kind": "jailbreak",
          "threshold": 0.50 },
        "pii_detector": {
          "kind": "pii",
          "threshold": 0.60,
          "pii_types_allowed": [
            "EMAIL_ADDRESS","GPE","AGE"] },
        "jira_intent": {
          "kind": "embedding",
          "threshold": 0.70,
          "candidates": [
            "create a jira issue",
            "open a bug ticket"] }
      },
      "signal_models": {
        "jb_guard": "llm-semantic-router/...",
        "pii_detector": "llm-semantic-router/..."
      }
    }
\end{lstlisting}

\subsection{Generated: Audit Trace Entry}

Listing~\ref{lst:audit_trace} shows a runtime audit
trace entry from a successful Jira routing decision.
Signal scores are from the real models in
Table~\ref{tab:signal_models}.

\begin{lstlisting}[style=python, caption={Audit trace
  entry with real model scores.},
  label=lst:audit_trace, float=t]
{
  "ts": 1711540200.123,
  "policy_version": "v2026.03.27",
  "source_hash": "a3f8c2d1",
  "tree": "outbound_delegation_gate",
  "branch": "allow_jira",
  "branch_idx": 3,
  "signals": {
    "jb_guard": 0.0000,
    "pii_detector": 0.0000,
    "jira_intent": 0.9983,
    "jira_intent_raw": 0.7465,
    "slack_intent": 0.0017,
    "slack_intent_raw": 0.1113,
    "authz_jira": True
  },
  "thresholds_crossed": {
    "jira_intent": "0.9983 > 0.5 (group)",
    "jira_intent_raw": "0.7465 > 0.70",
    "authz_jira": "True"
  }
}
\end{lstlisting}

\subsection{Generated: YANG Module (RFC~7950)}
\label{app:yang}

Listing~\ref{lst:yang_module} shows the YANG module
generated by the \texttt{EmitYANG} compiler target.
It defines signal types as YANG identities, signal
parameters as a keyed list with constrained types
(\texttt{decimal64\{fraction-digits 2; range 0..1\}}
for thresholds), and decision tree branches with
\texttt{ordered-by user} semantics.

\begin{lstlisting}[style=yaml, caption={Generated YANG
  module (truncated).},
  label=lst:yang_module, float=tp,
  basicstyle=\ttfamily\tiny]
module vllm-sr-policy {
  namespace "urn:vllm:semantic-router:policy";
  prefix vsr;
  revision 2026-03-27;

  identity signal-kind;
  identity authz { base signal-kind; }
  identity embedding { base signal-kind; }
  identity jailbreak { base signal-kind; }
  identity pii { base signal-kind; }

  container policy {
    leaf version { type string; }
    leaf source-hash {
      type string { length "8"; } }
    container signals {
      list signal {
        key "name";
        leaf name { type string; }
        leaf kind {
          type identityref {
            base signal-kind; } }
        leaf threshold {
          type decimal64 {
            fraction-digits 2;
            range "0.00..1.00"; } }
        leaf model { type string; }
        leaf-list candidates { type string; }
        leaf-list pii-types-allowed {
          type string; }
      }
    }
    container routing {
      list decision-tree {
        key "name";
        leaf name { type string; }
        list branch {
          key "priority";
          ordered-by user;
          leaf priority { type uint8; }
          leaf condition { type string; }
          leaf backend { type string; }
        }
      }
    }
    container network {
      list network-endpoint {
        key "name";
        leaf name { type string; }
        leaf host { type string; }
        leaf port { type uint16; }
        leaf skill { type string; }
      }
    }
  }
}
\end{lstlisting}

\subsection{Generated: NETCONF Edit-Config Payload}

Listing~\ref{lst:netconf_payload} shows the NETCONF
\texttt{<edit-config>} RPC generated from the same DSL
source.  It targets the candidate datastore and carries
the same \texttt{source-hash} as the YANG module and
Kubernetes artifacts.

\begin{lstlisting}[style=xml, caption={Generated NETCONF
  edit-config RPC (truncated).},
  label=lst:netconf_payload, float=tp,
  basicstyle=\ttfamily\tiny]
<?xml version="1.0" encoding="UTF-8"?>
<rpc xmlns="urn:ietf:params:xml:ns:netconf:base:1.0"
     message-id="1">
  <edit-config>
    <target><candidate/></target>
    <config>
      <policy xmlns=
        "urn:vllm:semantic-router:policy">
        <version>v2026.03.27</version>
        <source-hash>74f51a63</source-hash>
        <signals>
          <signal>
            <name>jb-guard</name>
            <kind>vsr:jailbreak</kind>
            <threshold>0.50</threshold>
            <model>llm-semantic-router/
              mmbert32k-jailbreak-detector
              -merged</model>
          </signal>
          <!-- ... pii, embedding signals ... -->
        </signals>
        <routing>
          <decision-tree>
            <name>outbound-gate</name>
            <branch>
              <priority>1</priority>
              <condition>jb-guard</condition>
              <backend>deny</backend>
            </branch>
            <!-- ... remaining branches ... -->
          </decision-tree>
        </routing>
      </policy>
    </config>
  </edit-config>
</rpc>
\end{lstlisting}

\subsection{Generated: OpenClaw Gateway Policy}
\label{app:openclaw}

Listing~\ref{lst:openclaw_config} shows the
\texttt{openclaw.json} fragment generated by the
compiler.  Two agents with different sandbox modes,
four channel bindings, and tool allow/deny lists
derived from \textsc{Network} endpoints are emitted
from the same DSL source.

\begin{lstlisting}[style=python, caption={Generated
  OpenClaw agent configuration (truncated).},
  label=lst:openclaw_config, float=tp,
  basicstyle=\ttfamily\tiny]
{
  "_dsl_metadata": {
    "source_hash": "ec1ff3ed",
    "policy_trees": ["safety_gate",
      "outbound_gate", "model_selector"]
  },
  "agents": {
    "list": [
      { "id": "dev-assistant",
        "model": "anthropic/claude-sonnet-4",
        "sandbox": { "mode": "non-main" },
        "tools": {
          "allow": ["bash", "read", "write",
            "skill:jira", "skill:slack"],
          "deny": ["browser", "canvas", "cron"]
        }
      },
      { "id": "ops-monitor",
        "model": "openai/gpt-4o-mini",
        "sandbox": { "mode": "all" },
        "tools": {
          "allow": ["bash", "read", "write",
            "skill:slack"],
          "deny": ["browser", "canvas", "cron"]
        }
      }
    ]
  },
  "bindings": [
    { "agentId": "dev-assistant",
      "match": { "channel": "slack",
                 "guildId": "T0123DEVTEAM" }},
    { "agentId": "ops-monitor",
      "match": { "channel": "telegram",
                 "peer": "ops-group-123" }}
  ],
  "channels": {
    "slack": { "dmPolicy": "pairing" },
    "telegram": { "dmPolicy": "allowlist" }
  },
  "session": {
    "sendPolicy": {
      "dsl_safety_gates": [
        { "signal": "jb_guard",
          "threshold": 0.50 },
        { "signal": "pii_detector",
          "threshold": 0.60,
          "pii_types_allowed":
            ["EMAIL_ADDRESS","GPE","AGE"] }
      ]
    }
  }
}
\end{lstlisting}

Listing~\ref{lst:openclaw_hook} shows the
\texttt{before\_tool\_call} plugin hook compiled from
the \textsc{outbound\_gate} decision tree.
The hook evaluates the same jailbreak and PII signals
used by LangGraph and protocol gate targets, at the
same thresholds, returning allow/deny with an audit
trace that includes the source hash.

\begin{lstlisting}[style=python, caption={Generated
  OpenClaw \texttt{before\_tool\_call} plugin hook
  (truncated).},
  label=lst:openclaw_hook, float=tp,
  basicstyle=\ttfamily\tiny]
// Auto-generated by DSL compiler
// (source_hash: ec1ff3ed)
// Decision tree: outbound_gate

import { evaluateSignal }
  from "./dsl-signal-evaluator";

export async function beforeToolCall({
  toolName, toolArgs, sessionKey, agentId,
}) {
  const text =
    `${toolName} ${JSON.stringify(toolArgs)}`;

  const jb_score = await evaluateSignal({
    kind: "jailbreak",
    model: "llm-semantic-router/" +
      "mmbert32k-jailbreak-detector-merged",
    input: text,
  });
  if (jb_score > 0.5) {
    return { action: "deny",
      signal: "jb_guard",
      score: jb_score,
      source_hash: "ec1ff3ed" };
  }

  const pii = await evaluateSignal({
    kind: "pii",
    model: "llm-semantic-router/" +
      "mmbert-pii-detector-merged",
    input: text,
    piiTypesAllowed: ["EMAIL_ADDRESS",
      "GPE", "AGE", "DATE_TIME"],
  });
  if (pii.score > 0.6) {
    return { action: "deny",
      signal: "pii_detector",
      score: pii.score,
      source_hash: "ec1ff3ed" };
  }

  return { action: "allow",
    tree: "outbound_gate",
    source_hash: "ec1ff3ed" };
}
\end{lstlisting}

\subsection{Agent Protocol Policy Gates}
\label{app:protocol_gates}

The same DSL-compiled policy evaluator runs at three
protocol boundaries.  Each gate uses the signal models
from Table~\ref{tab:signal_models} and applies the same
decision tree, with thresholds tuned for protocol message
context (embedding threshold lowered from 0.70 to 0.65
because protocol messages include tool names and
structured metadata that reduce cosine similarity).

\paragraph{MCP \texttt{tools/call} gate.}
The tool name is concatenated with the argument values
to form the evaluated text, providing semantic context
beyond raw JSON.  The gate catches jailbreak and PII
content in tool arguments while allowing legitimate
tool calls.

\paragraph{A2A \texttt{tasks/send} gate.}
Text parts from the A2A message are concatenated and
evaluated.  Safe delegations are allowed; injection
attempts are blocked by the same jailbreak signal.

\paragraph{Tool response gate.}
Evaluates tool output for indirect prompt injection
before passing it back to the LLM, forming a
defense-in-depth pattern with the input gate.

\subsection{Threshold Tuning Examples}

This section illustrates the tuning loop described in
\S\ref{sec:pillars:tune}.

\paragraph{LangGraph compilation.}  Initial thresholds
(jailbreak:~0.75, embedding:~0.78) were calibrated to
placeholder evaluators.  When replaced with the models in
Table~\ref{tab:signal_models}, several test cases failed:
the embedding model produces cosine similarities in the
0.70--0.75 range (not 0.85+), and the PII token
classifier flags geopolitical entities (GPE) that are
not sensitive PII.
The fix required changing only DSL declarations:
\texttt{threshold:~0.70} for embeddings,
\texttt{threshold:~0.50} for jailbreak (matching the
model's decision boundary), and adding \texttt{GPE},
\texttt{AGE}, \texttt{DATE\_TIME} to
\texttt{pii\_types\_allowed}.  No routing logic,
graph structure, or application code changed.
After recompilation, all test cases passed.

\paragraph{Protocol gate compilation.}  The same
thresholds from the orchestration target were applied to
protocol gates, where tests initially failed.  Protocol
messages (MCP JSON-RPC, A2A task payloads) include tool
names and structured metadata that reduce the embedding
model's cosine similarity compared to direct user queries.
The embedding threshold was lowered from 0.70 to~0.65
in the DSL declaration---a single-line change that
recompiled into all three protocol gates simultaneously.
After recompilation, all test cases passed.

\paragraph{Cross-target consistency.}
In both cases, threshold changes in DSL declarations
propagated to all compiler targets (LangGraph nodes,
Kubernetes ConfigMap, NETCONF payload, protocol gates)
on recompilation.  The source hash embedded in all
artifacts ensures version consistency across targets.

\section{LangGraph Implementation Notes}
\label{app:lg_internals}

This appendix documents the framework-specific details
underlying the compilation strategies of
\S\ref{sec:compilation}.  The observations are based on
the LangGraph v1.1.3 source code (modules
\texttt{graph/state.py}, \texttt{pregel/main.py},
\texttt{pregel/\_validate.py}).

\paragraph{State and channels.}
\texttt{StateGraph} is a builder whose state type is a
\texttt{TypedDict} or Pydantic model.  Each field becomes
a \emph{channel} with an optional reducer function.
Nodes are Python callables
$(state) \to \text{partial\_update}$.
Edges are static (\texttt{add\_edge}), conditional
(\texttt{add\_conditional\_edges(source, path\_fn)}),
or implicit via \texttt{Command(update=\ldots,
goto=\ldots)} returned from a node.

\paragraph{Compilation.}
Calling \texttt{.compile()} invokes \texttt{.validate()},
which checks: (i)~no orphan nodes, (ii)~all edge targets
exist, (iii)~\texttt{START} has outgoing edges.
It then builds a \texttt{Pregel} graph whose
execution model is \emph{channel-driven message passing}:
nodes fire when their trigger channels receive writes,
producing state updates and routing writes to
\texttt{branch:to:X} channels.

\paragraph{Conditional edge dispatch (Strategy~A).}
The conditional edge function is registered via
\texttt{add\_conditional\_edges}.  At runtime,
\texttt{BranchSpec.run()} reads fresh state (including
the signal evaluation node's writes) and invokes the
edge function.  The return type annotation
(\texttt{Literal["allow", "deny", \ldots]}) provides
build-time validation that all targets are registered
nodes.

\paragraph{Command dispatch (Strategy~B).}
When a node returns
\texttt{Command(update=\ldots, goto=\ldots)},
LangGraph's \texttt{\_control\_branch} processes the
\texttt{goto} field into channel writes, combining
state update, audit trace, and routing target in one
atomic operation.

\paragraph{Audit trace channel.}
The \texttt{audit\_trace} field uses an append reducer
(\texttt{Annotated[list, operator.add]}).  Downstream
nodes can append but not overwrite previous entries.
The checkpointer persists all state channels, making
the full audit trail available for compliance queries.

\end{document}